\documentclass{article}

\usepackage[final]{nips_2017}

\usepackage[utf8]{inputenc} 
\usepackage[T1]{fontenc}    
\usepackage{hyperref}       
\usepackage{url}            
\usepackage{booktabs}       
\usepackage{amsfonts}       
\usepackage{nicefrac}       
\usepackage{microtype}      

\usepackage{amsmath}
\usepackage{graphicx}
\usepackage{subfigure}
\usepackage{authblk}
\usepackage{amssymb}

\usepackage{listings}
\usepackage{color}
\definecolor{dkgreen}{rgb}{0,0.6,0}
\definecolor{gray}{rgb}{0.5,0.5,0.5}
\definecolor{mauve}{rgb}{0.58,0,0.82}
\title{Bayesian Paragraph Vectors}

\author[1]{\textbf{Geng Ji}}
\author[2]{\textbf{Robert Bamler}}
\author[1]{\textbf{Erik B. Sudderth}}
\author[2]{\textbf{Stephan Mandt}}
\affil[1]{Department of Computer Science, UC Irvine, \texttt{\{gji1,$\,$sudderth\}@uci.edu}}
\affil[2]{Disney Research, \texttt{firstname.lastname@disneyresearch.com}}

\begin{document}

\maketitle

\begin{abstract}
Word2vec \citep{mikolov2013distributed} has proven to be successful in natural language processing by capturing the semantic relationships between different words.
Built on top of single-word embeddings, paragraph vectors \citep{le2014distributed} find fixed-length representations for pieces of text with arbitrary lengths, such as documents, paragraphs, and sentences.
In this work, we propose a novel interpretation for neural-network-based paragraph vectors by developing an unsupervised generative model whose maximum likelihood solution corresponds to traditional paragraph vectors. 
This probabilistic formulation allows us to go beyond point estimates of parameters and to perform Bayesian posterior inference.
We find that the entropy of paragraph vectors decreases with the length of documents, and that information about posterior uncertainty improves performance in supervised learning tasks such as sentiment analysis and paraphrase detection.
\end{abstract}

\section{Introduction}

Paragraph vectors \citep{le2014distributed} are a recent method for embedding pieces of natural language text as fixed-length, real-valued vectors.
Extending the word2vec framework \citep{mikolov2013distributed}, paragraph vectors are typically presented as neural language models, and compute a single vector representation for each paragraph.
Unlike word embeddings, paragraph vectors are not shared across the entire corpus, but are instead local to each paragraph.
When interpreted as a latent variable, we expect them to have higher uncertainty when the paragraphs are short.

Recently, \citet{barkan2017bayesian} proposed a probabilistic view of word2vec that has motivated research on combining word2vec with other priors \citep{bamler2017dynamic}.
Inspired by this progress, we extend paragraph vectors to a probabilistic model.
Our model may be specified via modern inference tools like Edward \citep{tran2016edward}, which makes it easy to experiment with different inference algorithms.
The experiments in Sec.~\ref{sec:experiments} confirm the intuition that paragraph vectors have higher posterior uncertainty when paragraphs are short,
and we show that explicitly modeling this uncertainty improves performance in supervised prediction tasks.

\section{Related work}

Paragraph embeddings are built on top of word embeddings, a set of dimensionality reduction tools that map words from a large vocabulary to a dense vector representation.
Most word embedding methods learn a point estimate for each embedding vector \citep{mikolov_efficient_2013,mikolov2013distributed,mnih_learning_2013,goldberg_word2vec_2014,pennington2014glove}.
\citet{barkan2017bayesian} pointed out that the skip-gram model with negative sampling, also known as word2vec \citep{mikolov2013distributed}, admits a Bayesian interpretation.
The Bayesian skip-gram model allows uncertainty to be taken into account in a principled way, and lays the basis for our proposed Bayesian paragraph vector model.

Many tasks in natural language processing require fixed-length features for text passages of variable length, such as sentences, paragraphs, or documents (in this paper, we treat these three terms interchangeably).
Generalizing embeddings of single words, several methods have been proposed to find dense vector representations of paragraphs \citep{le2014distributed,kiros2015skip,wieting_paraphrase_2015,palangi2016deep,pagliardini_unsupervised_2017}.
Since paragraph embeddings are local to short pieces of text, we expect them to have high posterior uncertainty if the paragraphs are short.
In this work, we incorporate the idea of paragraph vectors \citep{le2014distributed} into the Bayesian skip-gram model in order to coherently infer the uncertainty associated with paragraph vectors.

\section{Method}
\label{sec:model}
\begin{figure}[t]
\center
	\begin{minipage}{0.46\linewidth}
	\centerline{\includegraphics[width=1.4in]{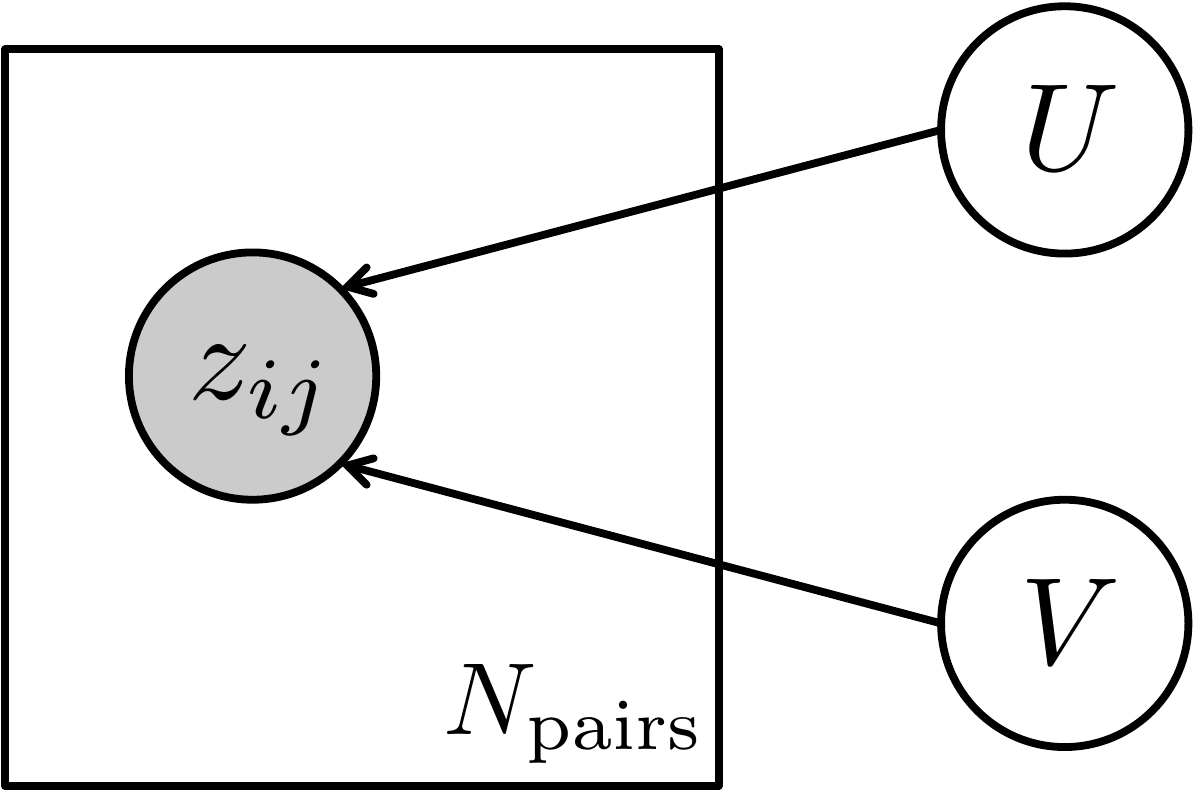}}
	\end{minipage}
	\hspace{.15cm}
	\begin{minipage}{0.46\linewidth}
	\centerline{\includegraphics[width=1.4in]{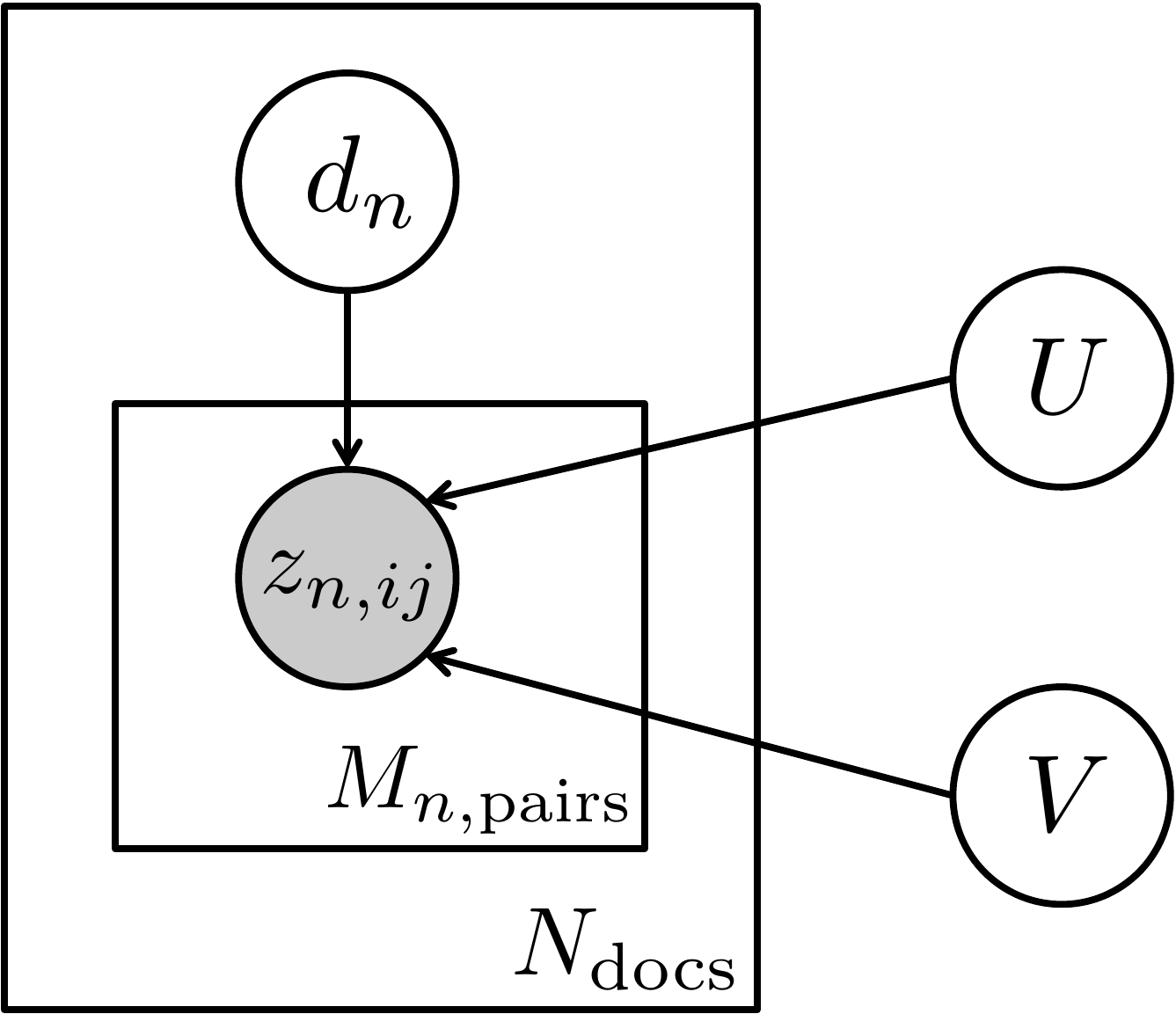}}
	\end{minipage}
    \vspace*{-0pt}
    \caption{\small Left: Bayesian word embeddings by \citet{barkan2017bayesian}. Right: Bayesian paragraph vectors.}
\label{fig:model}
\end{figure}

In Sec.~\ref{sec:bsg}, we summarize the Bayesian skip-gram model on which our model is based.
We then present our Bayesian paragraph model in Sec.~\ref{sec:bpv}, and discuss two inference methods in Sec.~\ref{sec:inference}.

\subsection{Bayesian skip-gram model}
\label{sec:bsg}

The Bayesian skip-gram model \citep{barkan2017bayesian} is a probabilistic interpretation of word2vec \citep{mikolov2013distributed}.
The left part of Figure~\ref{fig:model} shows the generative process.
For each word $i$ in the vocabulary, the model draws a latent word embedding vector $U_i \in \mathbb R^E$ and a latent context embedding vector $V_i \in \mathbb R^E$ from a Gaussian prior $\mathcal N(0, \lambda^2I)$.
Here, $E$ is the embedding dimension and $\lambda$ is a hyperparameter.
The model then constructs $N_\text{pairs}$ labeled pairs of words following a two-step process.
First, a proposal pair of words $(i,j)$ is drawn from a uniform distribution over the vocabulary.
Then, the model assigns to the proposal pair a binary label $z_{ij} \sim \text{Bern}(\sigma(U_i^\top V_j))$, where $\sigma(x)=1/(1+e^{-x})$ is the sigmoid function.
The pairs with label $z_{ij}=1$ form the so-called positive examples, and are assumed to correspond to occurrences of the word $i$ in the context of word $j$ somewhere in the corpus.
The so-called negative examples with label $z_{ij}=0$ do not correspond to any observation in the corpus.
When training the model, we resort to the heuristics proposed in \citep{mikolov2013distributed} to create artificial evidence for the negative examples (see Section~\ref{sec:bpv} below).

\subsection{Bayesian paragraph vectors}
\label{sec:bpv}

Bayesian paragraph vectors (BPV) are a direct extension of the Bayesian skip-gram model.
The right part of Figure~\ref{fig:model} shows the generative process.
In addition to global word and context embeddings $U$ and $V$, the model draws a paragraph vector $d_n \sim \mathcal{N}(0, \phi^2 I)$ for each of the $N_\text{docs}$ documents in the corpus.
Following \citet{le2014distributed}, we add $d_n$ to the context vector $V_j$ when we classify a given pair of words $(i,j)$ as a positive or a negative example.
Thus, the likelihood of a word pair $(i,j)$ in document $n$ to have label $z_{n,ij} \in\{0,1\}$ is
\begin{align}
    p(z_{n,ij} \mid U_i, V_j, d_n) = \sigma\big(U_i^\top (V_j + d_n)\big)^{z_{n,ij}}\,  \sigma\big(-U_i^\top (V_j + d_n)\big)^{1-z_{n,ij}}.
\end{align}

We collect evidence for the positive examples $\mathcal X^+_n$ in each document $n$ by forming pairs of words $(w_{n,t}, w_{n,t+\delta})$.
Here, $w_{n,t}$ is the word class of the $t$\textsuperscript{th} token, $t$ runs over all tokens in document $n$, $\delta$ runs from $-c$ to $c$ where $c$ is a small context window size, and we exclude $\delta=0$.
Negative examples are not observed in the corpus.
Following \citet{mikolov2013distributed}, we construct artificial evidence $\mathcal X^-_n$ for negative pairs by sampling from the noise distribution $P(i, j) \propto f(i) f(j)^\frac{3}{4}$, where $f$ is the empirical unigram frequency across the training corpus.
The log-likelihood of the entire data is thus
\begin{align}
    \label{eq:full-log-likelihood}
    \!\!\log p(\mathcal X^+, \mathcal X^-| U, V, d) = \sum_n \!\Bigg[
        \sum_{(i,j)\in\mathcal X^+_n}\!\!\!\! \log \sigma\big(U_i^\top (V_j + d_n)\big)
        + \!\!\!\!\!\sum_{(i,j)\in\mathcal X^-_n}\!\!\!\! \log \sigma\big(-U_i^\top (V_j + d_n)\big)
    \Bigg].\!
\end{align}
In the limit $d_n \to 0$, Eq.~\eqref{eq:full-log-likelihood} reduces to the negative loss function of word2vec.
BPV can be easily specified in Edward, a Python library for probabilistic modeling and inference \citep{tran2016edward}:
\begin{verbatim}
from edward.models import Bernoulli, Normal
U = Normal(loc=tf.zeros((W, E), dtype=tf.float32), scale=lam)
V = Normal(loc=tf.zeros((W, E), dtype=tf.float32), scale=lam)
d_n = Normal(loc=tf.zeros(E, dtype=tf.float32), scale=phi)
u_n = tf.nn.embedding_lookup(U, indices_n_I)
v_n = tf.nn.embedding_lookup(V, indices_n_J)
z_n = Bernoulli(logits=tf.reduce_sum(u_n * (v_n + d_n), axis=1))
\end{verbatim}

\subsection{MAP and black box variational inference}
\label{sec:inference}

The BPV model has global and local latent variables.
We expect the posterior of the global variables to be peaked, and therefore approximate the global word embedding matrices $U$ and $V$ via point estimates.
We expect a broader posterior distribution for the local paragraph vectors $d_n$.
Thus we use variational inference (VI) \citep{blei_variational_2016} to fit the posterior over $d_n$ with a fully factorized Gaussian distribution.
We split inference into two stages.
In the first stage, we point estimate all parameters.
In the second stage, we fix $U$ and $V$ and only perform VI for the paragraph vectors.

In the first stage, our goal is to train the global variables via stochastic gradient descent, where every minibatch contains a single document $n$ and a fixed set of negative examples $\mathcal X^-_n$.
We first maximize the joint probability $p(\mathcal X^+_n, \mathcal X^-_n, U,V,d_n)$ w.r.t the paragraph vector $d_n$.
As this local optimization is noise free, it converges quickly under a constant learning rate.
Then, we perform a single gradient step for the global variables $U$ and $V$.
This gradient is noisy due to the minibatch sampling and the stochastic generation of negative examples.
For this reason, a decreasing learning rate is used.
Finally, we reinitialize $d_n$ and proceed to the next document.
Optimizing $d_n$ in a nested loop before each update step saves memory since we only need to keep track of the document vectors one at a time.

In the second stage, we fit a variational distribution for the paragraph vectors while holding $U$ and $V$ fixed.
We use black box VI \citep{ranganath2014black} with reparameterization gradients \citep{kingma2014auto,rezende2014stochastic}, which is provided by the Edward library.
This time, we generate new negative examples in each update step to avoid overfitting.
The stochastic optimization is again performed with a decreasing learning rate.
We also perform a separate maximum a posteriori (MAP) estimate of the paragraph vectors to serve as the baseline for downstream classification tasks.

\section{Experiments}
\label{sec:experiments}

Paragraph vectors are often used as input features for supervised learning in natural language processing \citep{le2014distributed, kiros2015skip, palangi2016deep}.
In this section, we apply BPV to two binary classification tasks: sentiment analysis and paraphrase detection.
We find that the posterior uncertainty of BPV decreases as the length of paragraphs grows.
We also find that by concatenating the variational mean and standard deviation features inferred by VI, we improve  classification accuracy compared to MAP point estimates of paragraph embeddings.

\begin{figure}[t]
\center
	\begin{minipage}{0.48\linewidth}
	\centerline{\includegraphics[width=3in]{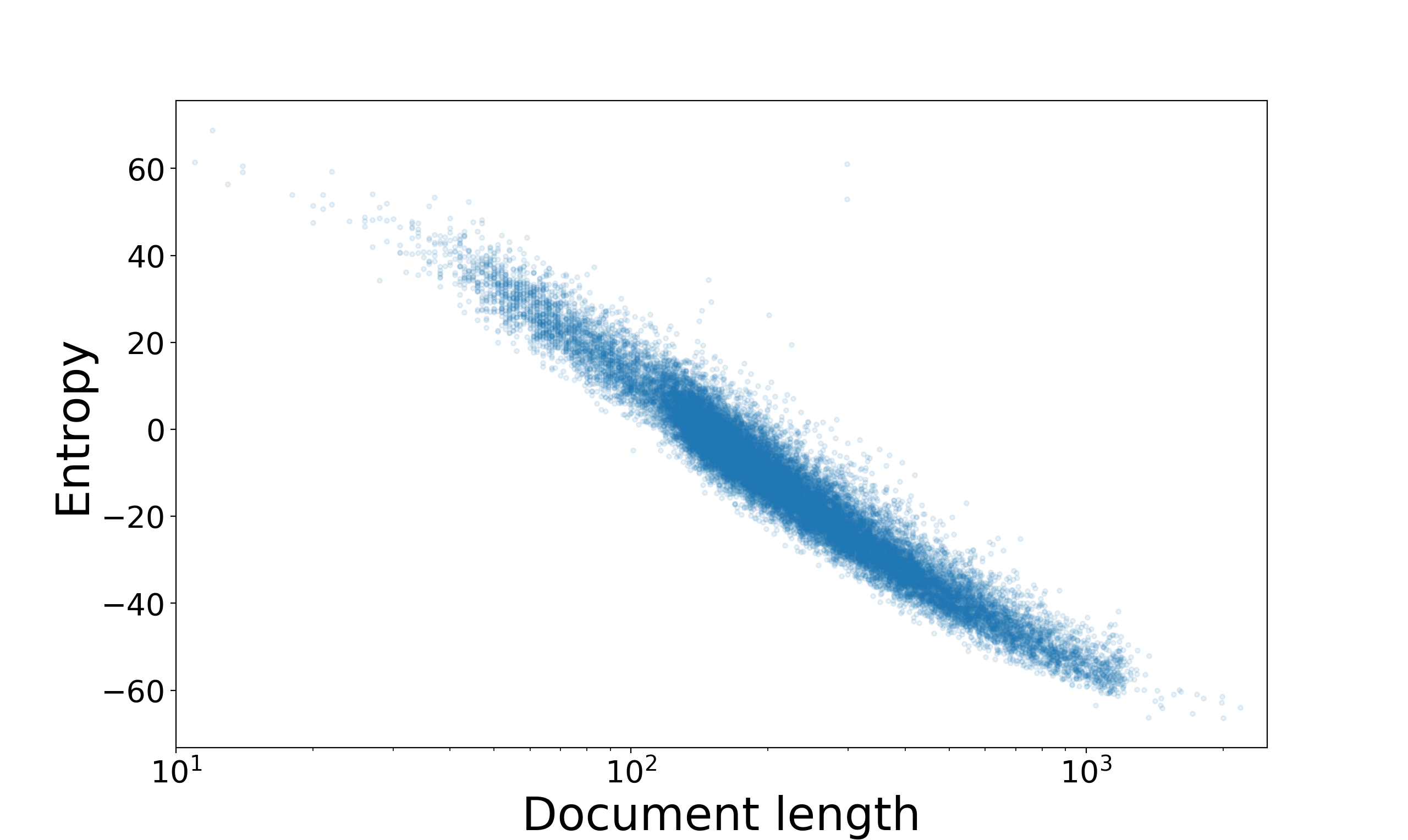}}
	\end{minipage}
	\hspace{.15cm}
	\begin{minipage}{0.48\linewidth}
	\centerline{\includegraphics[width=3in]{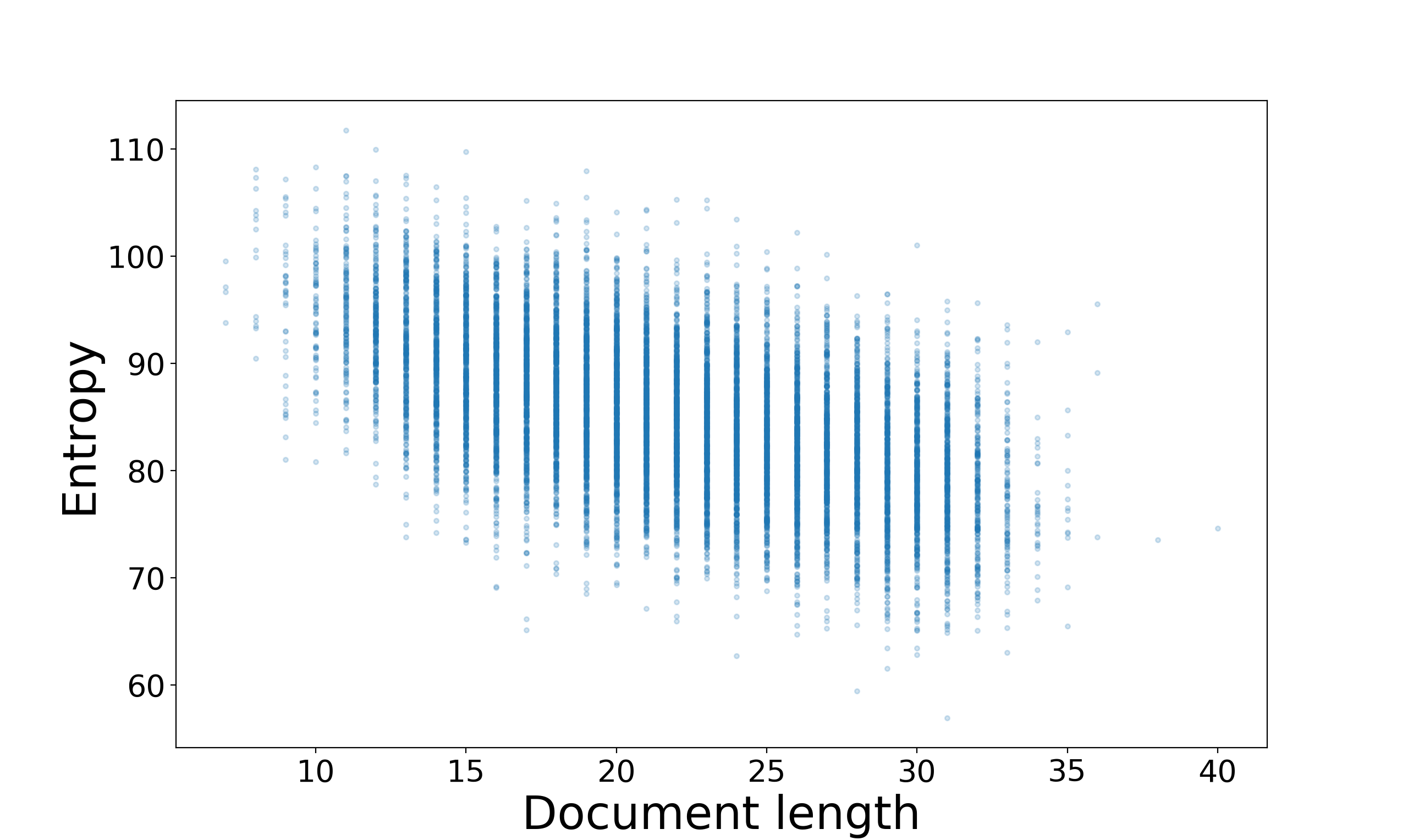}}
	\end{minipage}
    \vspace*{-0pt}
    \caption{\small Entropy of paragraph vectors as a function of the number of words in each document. Left: movie reviews in the IMDB dataset. Right: news clips in the MSR dataset. In the left plot, a log scale is used for the horizontal axis to account for the wide range of document lengths.}
\label{fig:plot}
\end{figure}

\subsection{Sentiment analysis}
We use the IMDB dataset \citep{maas2011learning} for sentiment analysis.
It contains 100k movie reviews, split into 75k training points (25k labeled, 50k unlabeled) and 25k labeled test points.
Positive and negative labels are balanced in both labeled subsets, and typical reviews consist of several sentences.

As our algorithm is unsupervised, we run the inference algorithms described in Sec.~\ref{sec:inference} using all the training data, and then train a logistic regression classifier using the paragraph vectors of the labeled training data only.
We use the most frequent 10k words as the vocabulary, and set the context window size $c=4$, the embedding dimension $E=100$, the hyperparameters for the prior $\lambda = \phi = 1$, and the number of negative examples per document equal to the average number of positive pairs of all the documents.
The feature vectors $x_n$ for the classifier are the point estimates of the paragraph vectors $d_n$ for MAP, and the concatenation of the variational mean and standard deviation of $d_n$ for VI.

Table~\ref{tab:accuracy} shows the test accuracy of the two inference methods.
VI outperforms MAP since it takes into account posterior uncertainty in paragraph embeddings.
Fig.~\ref{fig:plot} (left) shows the entropy of the paragraph vectors, computed using the posterior variance obtained from VI.
As the document length grows, the entropy decreases, which makes intuitive sense since longer reviews can be more specific.

\begin{table}[h]
  \caption{Classification accuracy of MAP and variational inference}
  \label{tab:accuracy}
  \centering
  \begin{tabular}{ccc}
    \toprule
    Task (dataset) & MAP & VI \\
    \midrule
    Sentiment analysis (IMDB)  & 86.9 & \textbf{87.0} \\
    Paraphrase detection (MSR) & 70.0 & \textbf{71.0} \\
    \bottomrule
  \end{tabular}
\end{table}

\subsection{Paraphrase detection}
We also test the discriminative power of BPV on the Microsoft Research Paraphrase Corpus \citep{dolan2004unsupervised}.
Each data point contains of two sentences extracted from news sources on the web, and the goal is to predict whether they are paraphrases of each other.
The training set contains 4076 sentence pairs in which 2753 are paraphrases, and the test set contains 1725 pairs among which 1147 are paraphrases.
We use the same hyperparameters as in the sentiment analysis task, except that we take all the words appearing more than once into the vocabulary because this dataset is much smaller.

After finding the paragraph vectors, we train the classifier by following \citet{kiros2015skip}, where features are constructed
by concatenating the component-wise product $x_n \cdot x_{n'}$ and the absolute difference $|x_n - x_{n'}|$ between each pair of features $x_n$ and $x_{n'}$.
The classification results in Table~\ref{tab:accuracy} show that VI again outperforms MAP.
The relationship between entropy and document length shown in Fig.~\ref{fig:plot} (right) is also similar to that of the IMDB dataset.

\section{Discussion}

We proposed Bayesian paragraph vectors, a generative model of paragraph embeddings. 
We treated the local latent variables of paragraph vectors in a Bayesian way because we expected high uncertainty, especially for short documents.
Our experiments confirmed this intuition, and showed that knowledge of the posterior uncertainty improves the performance of downstream supervised tasks.

In addition to MAP and VI, we experimented with Hamiltonian Monte Carlo (HMC) inference, but our preliminary results showed worse performance; we plan to investigate further.
A possible reason might be that we had to use a fixed set of negative examples for each document when generating HMC samples, which may result in overfitting to the noise.
Finally, we believe that more sophisticated models of document embeddings would also benefit from a Bayesian treatment of the local variables.

\newpage
{\small
\bibliographystyle{apa}
\bibliography{Reference}
}

\end{document}